\documentclass[runningheads,a4paper]{llncs}
\usepackage{multirow}
\usepackage[english]{babel}

\usepackage{amsmath}
\usepackage{graphicx}
\usepackage{multirow}
\usepackage{float}
\usepackage{caption}
\usepackage{subcaption}
\usepackage{hyperref}
\usepackage{booktabs}
\usepackage[table]{xcolor}

\graphicspath{{imgs/}}

\newcommand{\fig}[1]{Figure~\ref{fig:#1}}   

\newcommand{\tab}[1]{Table~\ref{tab:#1}}

\begin{document}

\title{Predicting Conversion of Mild Cognitive Impairments to Alzheimer's Disease and Exploring Impact of Neuroimaging}

\author{
    Yaroslav Shmulev \inst{1, 2}  \and 
    Mikhail Belyaev \inst{2, 1} \and
    the Alzheimer’s Disease Neuroimaging Initiative\thanks{Data used in preparation of this article were obtained from the Alzheimer’s Disease Neuroimaging Initiative (ADNI) database (\url{adni.loni.usc.edu}). As such, the investigators within the ADNI contributed to the design and implementation of ADNI and/or provided data but did not participate in analysis or writing of this report. A complete listing of ADNI investigators can be found at: \url{http://adni.loni.usc.edu/wp-content/uploads/how_to_apply/ADNI_Acknowledgement_List.pdf}}
}

\institute{
    Kharkevich Institute for Information Transmission Problems, Moscow, Russia \and Skolkovo Institute of Science and Technology, Moscow, Russia
}

\authorrunning{Y. Shmulev and M. Belyaev}
\tocauthor{Y. Shmulev and M. Belyaev}

\maketitle

\begin{abstract}
Nowadays, a lot of scientific efforts are concentrated on the diagnosis of Alzheimer’s Disease (AD) applying deep learning methods to neuroimaging data. Even for 2017, there were published more than a hundred papers dedicated to AD diagnosis, whereas only a few works considered a problem of mild cognitive impairments (MCI) conversion to AD. However, the conversion prediction is an important problem as approximately 15\% of patients with MCI converges to AD every year.  In the current work, we are focusing on the conversion prediction using brain Magnetic Resonance Imaging and clinical data, such as demographics, cognitive assessments, genetic, and biochemical markers. First of all, we applied state-of-the-art deep learning algorithms on the neuroimaging data and compared these results with two machine learning algorithms that we fit using the clinical data. As a result, the models trained on the clinical data outperform the deep learning algorithms applied to the MR images. To explore the impact of neuroimaging further, we trained a deep feed-forward embedding using similarity learning with Histogram loss on all available MRIs and obtained 64-dimensional vector representation of neuroimaging data. The use of learned representation from the deep embedding allowed to increase the quality of prediction based on the neuroimaging. Finally, the current results on this dataset show that the neuroimaging does affect conversion prediction, however, cannot noticeably increase the quality of the prediction. The best results of predicting MCI-to-AD conversion are provided by XGBoost algorithm trained on the clinical and embedding data. The resulting accuracy is $\textrm{ACC} = 0.76 \pm 0.01$ and the area under the ROC curve -- $\textrm{ROC AUC} = 0.86 \pm 0.01$. 

\end{abstract}

\section{Introduction}
    Alzheimer's Disease is irreversible progressive brain disorder mostly occurring in the middle or late life. At the same time, there is a transitional phase between the normal aging and dementia symptoms called mild cognitive impairment (MCI). People with MCI are at increased risk of AD development -- approximately 15\% of them converge to dementia every year. That's why the early diagnosis of Alzheimer's Disease would allow patients to take preventive measures to temporarily slow the disease progression \cite{MCIprogression}. 
    
    Neuroimaging is a variety of methods and technologies that reveal the structure and functions of the brain. It includes Computer Tomography (CT), structural and functional Magnetic Resonance Imaging (sMRI and fMRI respectively) and etc. With a growth of deep learning applications in data analysis, neuroimaging is extensively used in many medical tasks such as image segmentation \cite{Segment}, diagnosis classification \cite{DEEPref1} and prediction of disease progression \cite{conv1}. 
    
    In the recent years, there were published a vast number of papers dedicated to classification of healthy controls from AD using deep learning approach applied to neuroimaging. However, only a few works considered predicting conversion of MCI to AD (\cite{conv1}, \cite{conv2}, \cite{conv3}), which is a more complicated and clinically relevant problem. To classify stable and converged MCI the authors of \cite{conv2} used different clinical biomarkers and complex feature maps extracted from neuroimaging.
    This method inherits the main drawbacks from manual feature extraction procedure.
    Cheng et. al in \cite{conv1} consider the joint multi-domain learning for early diagnosis of AD to boost the learning performance. 
    In this work, we are focusing on the conversion prediction using clinical and neuroimaging data. In addition, we want to explore the individual impact of different data types on the prediction performance for different prediction intervals. Finally, we obtain a low-dimensional representation of high-dimensional MR brain images from a deep feed-forward embedding that is trained on the whole ADNI cohort.  
\section{Data} \label{sec_data}

    In this work, we use data obtained from the Alzheimer's Disease Neuroimaging Initiative (ADNI) database \cite{ADNI}. We choose patients that are diagnosed as normal controls (NC), mild cognitive impairments (MCI), and Alzheimer's Disease (AD). For each patient, we take visits for which both MR images and clinical data are available. The total number of available data samples is 8809.
    
    \textbf{Clinical data.}
    ADNI provides clinical information about each subject including recruitment, demographics, physical examinations, and cognitive assessment data. We add genetic and biospecimen data (cerebrospinal fluid concentration, blood, and urine) to the clinical dataset. The full list of attributes is available on the official ADNI website \cite{ADNI}.

    \textbf{Neuroimaging data.}
    For the analysis, we take structural T1-weighted Magnetic Resonance Imaging (MRI), since they are available for all patients and for the most their visits. We fetch preprocessed images from ADNI database with the following preprocessing pipeline descriptions: \textbf{``MPR; GradWarp; B1 Correction; N3; Scaled''} and \textbf{``MT1; GradWarp; N3m''}. These images have different shapes and orientations and contain skull and other organs that might spoil a predictive performance. Thus, we apply the following preprocessing pipeline for the collected neuroimaging dataset.
    For the Brain extraction \cite{BrainExtract} and N4 bias correction \cite{N4} steps, we run ANTs Cortical Thickness Pipeline \cite{ANTs} for all available MR images. Then, we apply an affine transform, so that all brain images have the same orientation - \textbf{RAS} (Left - Right, Posterior - Anterior, Superior - Inferior). After the brain extraction step, the MRIs contain a lot of black voxels around the brain. We crop all images to the maximal extracted brain size, which is computed beforehand. Ultimately, all MR images in the dataset have a size of $(150, 208, 173)$. To increase a batch size that can be fitted to the Graphics Processing Unit (GPU), we downsample the dataset with the factor of 2 for each dimension, so that the resulting shape is $(75, 104, 87)$.

    \textbf{Conversion dataset.}
    To predict the MCI-to-AD conversion, we need to remove patients that are normal controls (NC) or have Alzheimer's Disease from the screening visit.

    For the stable MCI, we consider participants that have not converged to AD for the known time-period. We also drop several last visits that are inside the prediction horizon, since the future for that patients is not known and they may converge to AD in the next visits.

    For the converged MCI, we select participants that were diagnosed as MCI in earlier sessions and as AD later. We take visits that are within five-year prediction interval. The total number of stable and converged patients are 532 and 327 correspondingly. The number of samples for the two classes are 1764 and 1016.
\section{Method}
    \subsection{Clinical data}
        For the classification based on clinical data, we use two machine learning algorithms: Logistic Regression and XGBoost. The first one is a linear method which is widely used in many practical applications because of its good interpretability and relative simplicity. The second method is an efficient implementation of gradient boosting on decision trees, which is a powerful machine learning algorithm that can catch nonlinear patterns in data \cite{XGBoost}.
    
    \subsection{Neuroimaging data}
        Convolutional Neural Networks (CNN) have recently made a great breakthrough in the image classification and recognition tasks. Deep CNNs automatically extract and combine from low- to high-level features from images and estimate target values in the end-to-end fashion. 
        In this work, we use two deep architectures: VGG \cite{VGG} and ResNet \cite{ResNet}, that showed state-of-the-art performance in ImageNet classification challenge in 2014 and 2015 correspondingly. We generalize these architectures to the three-dimensional input size of MR images in the same way as was proposed in \cite{DEEPref2}. 
        
        \textbf{VoxCNN.}
        The VGG-like network consists of ten 3D convolutional blocks, each of which consists of three 3D-convolutional layers with 3x3x3 filter sizes, batch normalization and ReLU nonlinearity. Then, we use max pooling layer with 2x2x2 kernel size to reduce the size of data propagated through the network. At the end of the net, there are three fully-connected layers with batch normalization and dropout layers in-between. For the experiments, we used the probability $p = 0.7$ of a neuron to be turned off. After the last fully-connected layer, there is softmax activation function to compute probabilities for each class.

        \textbf{ResNet3D}.
        For the ResNet-like architecture, we use 6 residual blocks, each of which represents a sum of identity mapping and a stack of two convolutions with 3x3x3 filter size and 64 or 128 filters, batch normalization and ReLU. The convolutional layers from the standard ResNet are replaced with 3D ConvBlocks in the same way as we did for VoxCNN. We reduce the spacial size using three convolutions with strides 2x2x2 before the residual blocks and one maximum pooling layer with 5x5x5 kernel size before the fully-connected layer. Dropout with $p = 0.7$ and batch normalization are also used after the first fully-connected layer. At the end of the network, there is a second fully-connected layer with softmax activation to produce output probabilities.
\section{Experiments}
    \subsection{Setup}
        For the experiments with conversion prediction based on the neuroimaging data, we minimize a weighted binary cross-entropy loss function with balanced class weights.
        We use Nesterov momentum optimizer with initial learning rate $10^{-3}$ and scheduling learning rate policy: we decrease the learning rate ten-fold after 30 and 50 epochs. The batch sizes for ResNet3D and VoxCNN are 128 and 512 correspondingly. These numbers are chosen so that the full batch can be fitted to the GPU. The total number of epochs is 70.

    \subsection{Validation}
        To assess the classification performance more accurately, we run 5-fold group cross-validation with five different folds. As a group label, we use participant's IDs to prevent the cases when different scans of one patient are simultaneously in train and test sets.
        
        For neuroimaging data, on each step of cross-validation procedure, we train a separate neural network on a train set, use a validation set for early stopping and changing learning rate, and test the network model on a hold-out subset. 
        
        For hyperparameter tuning of Logistic Regression and XGBoost methods, another nested group cross-validation procedure is used.
        
        We report the following metrics: accuracy, the area under the receiver operating characteristic curve (ROC AUC), sensitivity, specificity, and average precision.
    
    \subsection{Conversion prediction}
        There are several ways how the disease progression problem can be formulated. In this work, we use binary classification setting to predict the fact of conversion within a five year interval: \textbf{class 0} - stable MCI, \textbf{class 1} - converged MCI.
        In other words, given a participant's visit, we would like to answer the question, whether an individual will converge to AD within the considered interval or not.
        
    \subsection{Embedding Learning}
\begin{figure}[ht]
            \centering
            \begin{subfigure}[t]{0.33\textwidth}
                \includegraphics[width=\textwidth]{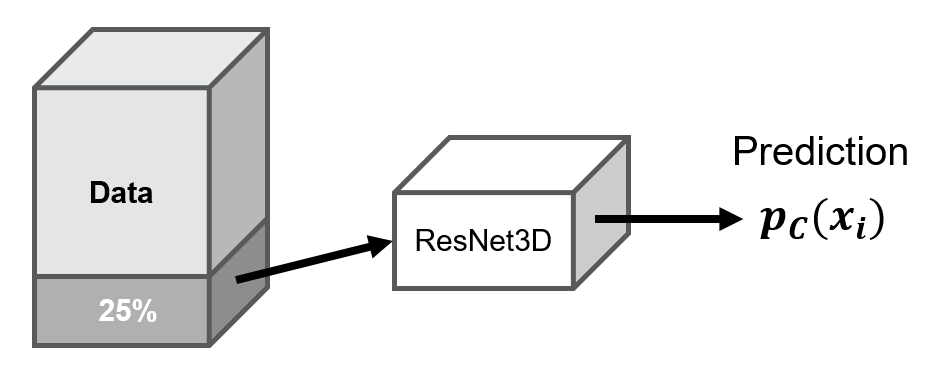}
                \caption{Current approach}
                \label{fig:embedding_current}
            \end{subfigure}
            \begin{subfigure}[t]{0.55\textwidth}
                \includegraphics[width=\textwidth]{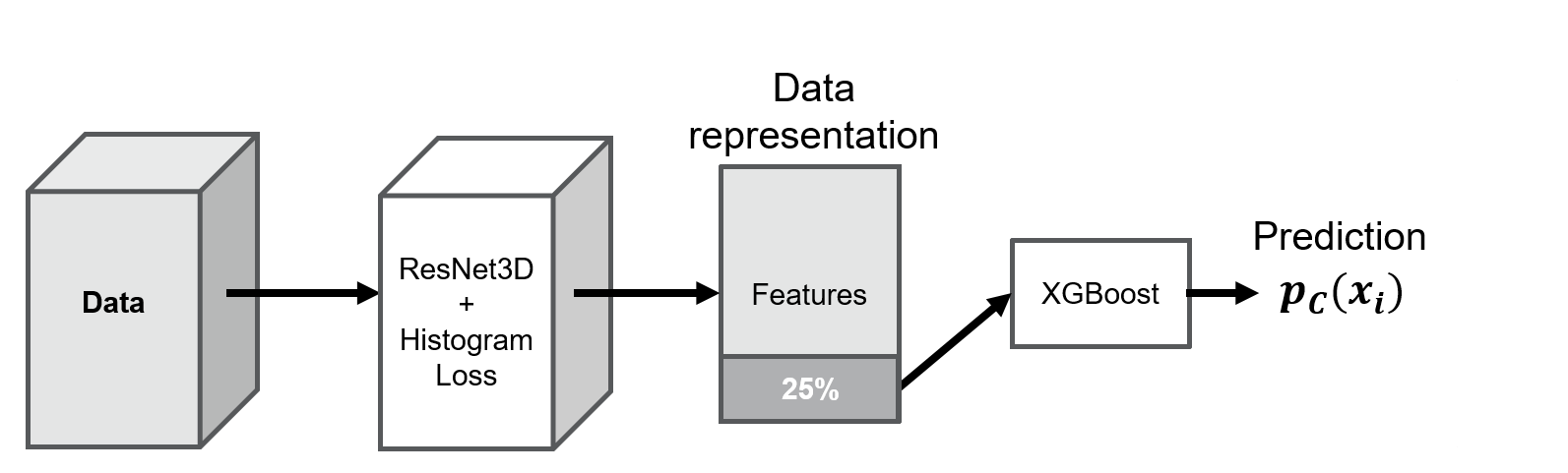}
                \caption{Deep embedding approach}
                \label{fig:embedding_deep}
            \end{subfigure}
            \caption[Two approaches for conversion prediction task]{Two approaches for conversion prediction task. a) In the current approach, only 25\% of all available MR images are used for the conversion prediction. b) The embedding is trained on the whole MRI dataset and, then, used for feature extraction. We use the extracted features for the conversion prediction task.}
            \label{fig:embedding}
        \end{figure}

        For conversion prediction, we used only 25\% of all available MR images. To make use of all available data, we learn a deep feed-forward embedding on the whole neuroimaging dataset and, then, use it as a fixed feature extractor. We exploit the extracted features for conversion prediction task, as shown in \fig{embedding}.
        
        Deep embedding is an approach when complex high-dimensional input data are mapped into a smaller size semantic subspace preserving the most relevant information about the data. Generally, the mapping is learned from a large amount of supervised data. During the training process, semantically related samples are getting closer than semantically unrelated ones in the semantic subspace. To learn deep feed-forward embedding we use ResNet3D architecture up to the last fully connected layer. We add a fully connected layer with 64 output units and L2-normalization layer to the network. We use Histogram loss proposed in \cite{HLoss} as a training criterion, which is parameter-free batch loss function that firstly estimates two distributions of distances between matching and non-matching pairs and, secondly, computes the overlap between these two distributions.
        Once the deep embedding is trained, we use it to extract the embedded features: all images from the conversion dataset are propagated through the network and 64-dimensional vector representations are obtained. 
\section{Results}
    The results of the conversion prediction are shown in \tab{conv_pred_results}.
    For the considered interval, the quality of prediction based on clinical data is significantly higher than we achieved using neuroimaging data. On the neuroimaging, two network architectures provide comparative results for all experiments, although ResNet3D slightly outperforms VoxCNN. For the clinical data, the performance of XGBoost is slightly better than one of the Logistic Regression model.

    The results also show that the use of learned deep embedding helps increase the quality of prediction based on MR images, although it is still worse than one on the clinical data.

    To investigate whether the neuroimaging data can add some new relevant information to the clinical data and, thereby, improve the prediction, we include extracted features from the embedding to the clinical ones.
    As can be seen from the results, the quality of prediction using clinical and embedding data is slightly higher than for clinical data, although still, it is the same within the standard deviation. 
    
    \begin{table}[ht]
    \centering
     \begin{tabular}{|l|ccccc|} 
     \hline
     \textbf{Data / Method} & ACC & ROC AUC & AV PREC & SENS & SPEC \\ [0.5ex]
     \hline
     \hline
     Clinical data / Log Reg & $.76 \pm .01$ & $.85 \pm .01$ & $.73 \pm .05$ & $.80 \pm .03$ & $.74 \pm .02$ \\ 
     Clinical data / XGBoost  & $.76 \pm .01$ & $.85 \pm .01$ & $.73 \pm .03$ & $.76 \pm .02$ & $.77 \pm .01$ \\  [0.5ex]
     \hline
    Neuroimaging / VoxCNN & $.61 \pm .02$ & $.70 \pm .03$ & $.52 \pm .05$ & $.70 \pm .04$ & $.56 \pm .02$ \\ 
     Neuroimaging / ResNet3D &  $.62 \pm .01$ & $.70 \pm .02$ & $.53 \pm .02$ & $.75 \pm .03$ & $.54 \pm .01$ \\  [0.5ex]
     \hline
     Embedding / Log Reg  & $.69 \pm .01$ & $.71 \pm .01$ & $.54 \pm .03$ & $.60 \pm .01$ & $.75 \pm .03$ \\ 
     Embedding / XGBoost &  $.67 \pm .02$ & $.73 \pm .01$ & $.57 \pm .02$ & $.70 \pm .02$ & $.65 \pm .05$ \\  [0.5ex]
     \hline
     Clinic. + Embed. / Log Reg  & $.76 \pm .02$ & $.86 \pm .02$ & $.73 \pm .03$ & $.84 \pm .02$ & $.72 \pm .03$ \\ 
     Clinic. + Embed. / XGBoost &  $.76 \pm .01$ & $.86 \pm .01$ & $.73 \pm .02$ & $.88 \pm .03$ & $.70 \pm .03$ \\  [0.5ex]
     \hline
     \end{tabular}
    \caption{Conversion prediction results}
    \label{tab:conv_pred_results}
    \end{table}

\begin{figure}[ht!]
        \centering
        \begin{subfigure}[t]{0.33\textwidth}
            \includegraphics[width=\textwidth]{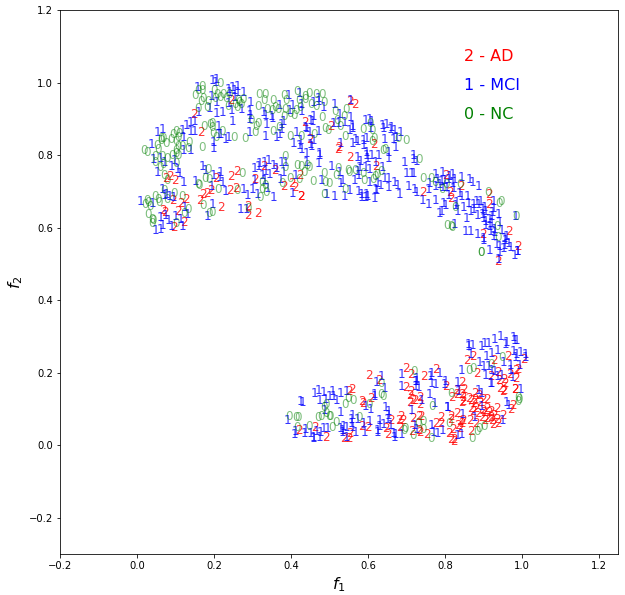}
            \caption{Clusters}
            \label{fig:embedded_clusters}
        \end{subfigure}
        \quad
        \begin{subfigure}[t]{0.33\textwidth}
            \includegraphics[width=\textwidth]{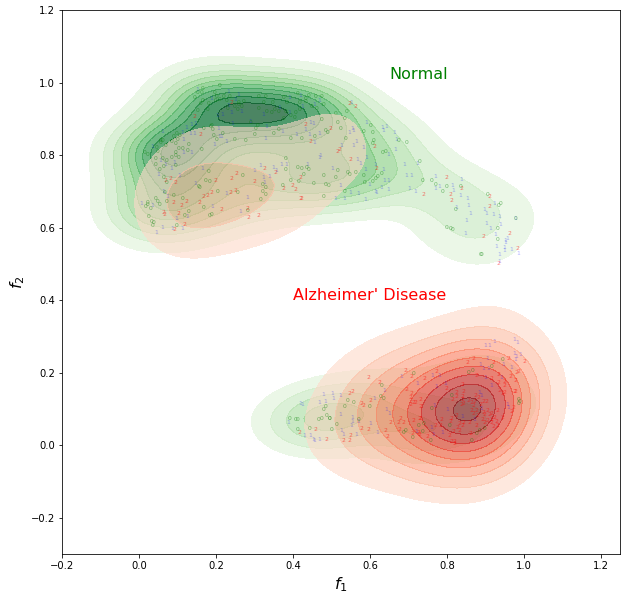}
            \caption{NC / AD density estimation}
            \label{fig:kde_ad_nc}
        \end{subfigure}
        \\
        \centering
        \begin{subfigure}[t]{0.33\textwidth}
            \includegraphics[width=\textwidth]{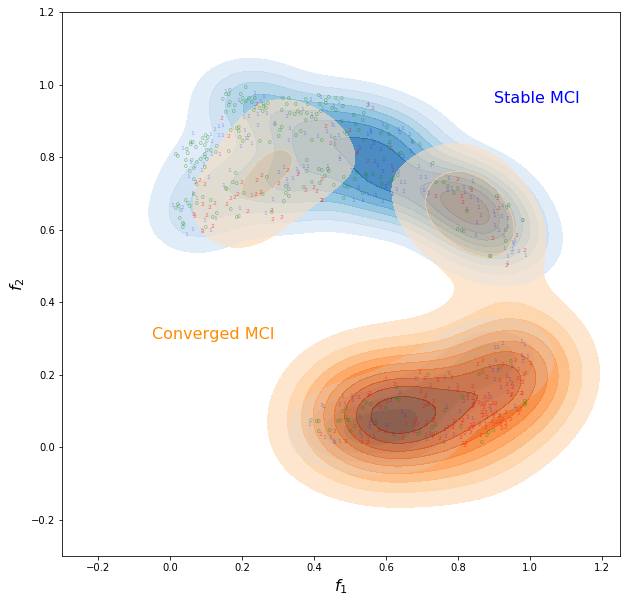}
            \caption{sMCI / cMCI density estimation}
            \label{fig:kde_mci}
        \end{subfigure}
        \quad
        \begin{subfigure}[t]{0.33\textwidth}
            \includegraphics[width=\textwidth]{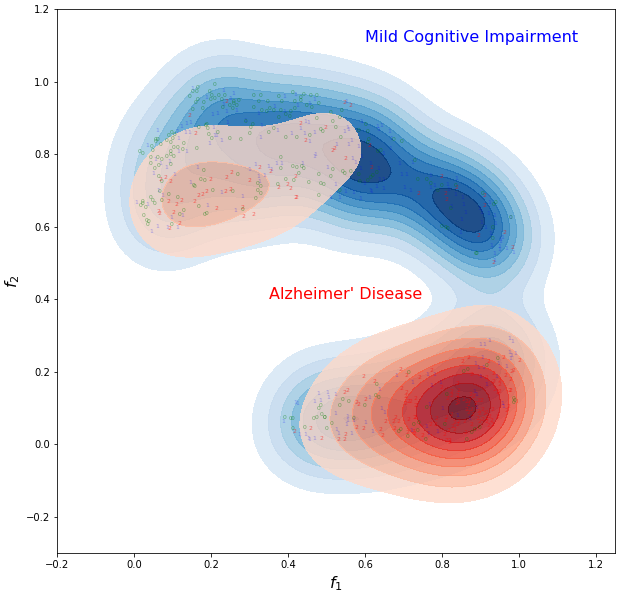}
            \caption{MCI / AD density estimation}
            \label{fig:kde_ad_mci}
        \end{subfigure}
        \caption[Embedding visualization]{Embedding visualization: a) Clusters in the embedded space, b) Kernel Density Estimation (KDE) of NC and AD distributions, c) KDE of stable and converged MCI distributions, d) KDE of MCI and AD distributions.}
        \label{fig:embed_viz}
    \end{figure}

    \fig{embed_viz} shows the resulting representation from the learned deep embedding on a hold-out set. We applied T-SNE algorithm proposed in \cite{tsne} to map our 64-dimensional feature vectors into 2-dimensional ones. In \fig{embedded_clusters}, there are three clusters, each of which corresponds to one of the diagnoses: NC, MCI, or AD. From \fig{kde_ad_nc} and \fig{kde_ad_mci} can be seen that the separation between NC and AD is better than separation between MCI and AD.
    
    The next observation from \fig{embedded_clusters} is that MCI cluster is spread between normal controls (NC) and Alzheimer's Disease (AD). Since we know which MCI patient will converge to AD (cMCI) and which will not (sMCI), we plot the densities of stable and converged MCI. \fig{kde_mci} shows that these two groups of MCIs are quite good separated in the embedded space. The main mass of converged MCIs is closer to the AD cluster, whereas the stable MCIs are closer to the normal controls.
\section{Conclusion}

    In this work, a problem of conversion prediction from mild cognitive impairment (MCI) to Alzheimer's Disease (AD) was considered. We collected, preprocessed and analyzed the clinical and neuroimaging data. We applied the state-of-the-art methods for image classification on the neuroimaging data and compared the quality of classification with the several machine learning methods trained on the clinical data. The results of the experiments showed that the clinical data allow to obtain a better prediction quality than the neuroimaging and these models can be used for conversion prediction task.

    We enhanced the performance on the neuroimaging data by training a deep feed-forward embedding. The embedding increased the quality of the forecast, however, it is still worse than the clinical data yield. We further investigate the question whether the neuroimaging is able to add some new information for conversion prediction or not. According to the results on the current dataset, neuroimaging does have an effect on the conversion prediction, however, it cannot noticeably increase the quality of the prediction when clinical data are used.
    
\section*{Acknowledgements}
The results have been obtained under the support of the Russian Science Foundation grant 17-11-0139.

\bibliographystyle{splncs03}
\bibliography{main}

\end{document}